# Comparison of the Bayesian and Randomised Decision Tree Ensembles within an Uncertainty Envelope Technique

Vitaly Schetinin, Jonathan E. Fieldsend, Derek Partridge, Wojtek J. Krzanowski, Richard M. Everson, Trevor C. Bailey, and Adolfo Hernandez

School of Engineering, Computer Science and Mathematics, University of Exeter, EX4 4QF, UK

**Abstract:** Multiple Classifier Systems (MCSs) allow evaluation of the uncertainty of classification outcomes that is of crucial importance for safety critical applications. The uncertainty of classification is determined by a trade-off between the amount of data available for training, the classifier diversity and the required performance. The interpretability of MCSs can also give useful information for experts responsible for making reliable classifications. For this reason Decision Trees (DTs) seem to be attractive classification models for experts. The required diversity of MCSs exploiting such classification models can be achieved by using two techniques, the Bayesian model averaging and the randomised DT ensemble. Both techniques have revealed promising results when applied to real-world problems. In this paper we experimentally compare the classification uncertainty of the Bayesian model averaging with a restarting strategy and the randomised DT ensemble on a synthetic dataset and some domain problems commonly used in the machine learning community. To make the Bayesian DT averaging feasible, we use a Markov Chain Monte Carlo technique. The classification uncertainty is evaluated within an Uncertainty Envelope technique dealing with the class posterior distribution and a given confidence probability. Exploring a full posterior distribution, this technique produces realistic estimates which can be easily interpreted in statistical terms. In our experiments we found out that the Bayesian DTs are superior to the randomised DT ensembles within the Uncertainty Envelope technique.

## 1. Introduction

The uncertainty of Multiple Classifier Systems (MCSs) used for safety-critical applications, such as medical diagnostics, air traffic control, *etc*., is of crucial importance. In general, uncertainty is a triple trade-off between the amount of data available for training, the classifier diversity and the classification accuracy [1-5]. The interpretability of classifiers can also give useful information to experts responsible for making reliable classifications. For this reason Decision Trees (DTs) seem to be attractive classification models for experts [1, 4].

The required diversity of a MCS can be achieved by using two approaches, an averaging technique based on Bayesian Markov Chain Monte Carlo (MCMC) search methodology [1-3] and an ensemble technique [4, 5]. Both techniques have revealed promising results when applied to some real-world problems [1-5].

The main idea of using DT classification models is to recursively partition data points in an axis-parallel manner. Such models provide natural feature selection and uncover the most important features for the classification. The resultant DT classification models are easily interpretable by users.

By definition, DTs consist of splitting and terminal nodes, which are also known as tree leaves. DTs are said to be binary if the splitting nodes ask a specific question and then divide the data points into two disjoint subsets, say the left or the right branch. The terminal node assigns all data points falling in that node to a class of majority of the training data points reached this terminal node. Within a Bayesian framework, the class posterior distribution is calculated for each terminal node [1-3].

Breiman *et al.* [1] have given the Bayesian generalisation of tree models required to evaluate the posterior distribution. Recently Denison *et al.* [3] have suggested a MCMC technique for evaluating the posterior distribution of DTs. This technique performs stochastic sampling from the posterior distribution.

In this paper we compare the classification uncertainty of the Bayesian DT technique with a restarting strategy, and the randomised DT ensemble technique, on a synthetic dataset and some domain problems commonly used in the machine learning community. The classification uncertainty is evaluated within an Uncertainty Envelope dealing with the class posterior distribution and a given confidence probability as described in [6]. By estimating the consistency of MCS outputs on the given data, the Uncertainty Envelope produces estimates of the classification uncertainty which can be easily interpreted in statistical terms. Using this evaluation technique in our experiments, we found that the Bayesian DT technique is superior to the randomised DT ensemble technique

In sections 2 and 3 we describe the randomised and Bayesian DT techniques which are used in our experiments. Then in section 4 we briefly describe the Uncertainty Envelope technique used for comparison of the uncertainty of the two classification techniques. The experimental results are presented in section 5, and section 6 concludes the paper.

## 2. The Bayesian Decision Tree Technique

In this section we present the Bayesian DT technique based on MCMC search methodology. We then discuss the difficulties of searching the posterior distribution, which can be resolved within the restarting strategy of the MCMC technique.

### 2.1. Bayesian Decision Trees

In general, the predictive distribution we are interested in is written as an integral over parameters $\boldsymbol{\theta}$ of the classification model:

$$p(y \mid \mathbf{x}, \mathbf{D}) = \int_{\theta} p(y \mid \mathbf{x}, \boldsymbol{\theta}, \mathbf{D}) p(\boldsymbol{\theta} \mid \mathbf{D}) d\boldsymbol{\theta} \tag{1}$$

where $y$ is the predicted class $(1, \ldots, C)$, $\mathbf{x} = (x_1, \ldots, x_m)$ is the $m$-dimensional input vector, and $\mathbf{D}$ denotes the given training data.

The integral (1) can be analytically calculated only in simple cases. In practice, part of the integrand in (1), which is the posterior density of $\boldsymbol{\theta}$ conditioned on the data $\mathbf{D}$, $p(\boldsymbol{\theta} \mid \mathbf{D})$, cannot usually be evaluated. However if values $\boldsymbol{\theta}^{(1)}, \ldots, \boldsymbol{\theta}^{(N)}$ are drawn from the posterior distribution $p(\boldsymbol{\theta} \mid \mathbf{D})$, we can write

$$p(y \mid \mathbf{x}, \mathbf{D}) \approx \sum_{i=1}^{N} p(y \mid \mathbf{x}, \boldsymbol{\theta}^{(i)}, \mathbf{D}) p(\boldsymbol{\theta}^{(i)} \mid \mathbf{D}) = \frac{1}{N} \sum_{i=1}^{N} p(y \mid \mathbf{x}, \boldsymbol{\theta}^{(i)}, \mathbf{D}) \cdot \qquad (2)$$

This is the basis of the MCMC technique for approximating integrals [3]. To perform the approximation, we need to generate random samples from $p(\boldsymbol{\theta} \mid \mathbf{D})$ by running a Markov Chain until it has converged to a stationary distribution. After this we can draw samples from this Markov Chain and calculate the predictive posterior density (2).

Let us now define a classification problem presented by data $(\mathbf{x}_i, y_i)$, $i = 1, \ldots, n$, where $n$ is the number of data points and $y_i \in \{1, \ldots, C\}$ is a categorical response. Using DTs for classification, we need to determine the probability $\varphi_{ij}$ with which a datum $\mathbf{x}$ is assigned by terminal node $i = 1, \ldots, k$ to the $j$th class, where $k$ is the number of terminal nodes in the DT. Initially we can assign a $(C - 1)$-dimensional Dirichlet prior for each terminal node so that $p(\boldsymbol{\varphi}_i \mid \boldsymbol{\theta}) = \mathrm{Di}_{C-1}(\boldsymbol{\varphi}_i \mid \boldsymbol{\alpha})$, where $\boldsymbol{\varphi}_i = (\varphi_{i1}, \ldots, \varphi_{iC})$, $\boldsymbol{\theta}$ is the vector of DT parameters, and $\boldsymbol{\alpha} = (\alpha_1, \ldots, \alpha_C)$ is a prior vector of constants given for all the classes.

The DT parameters are defined as $\boldsymbol{\theta} = (s_i^{pos}, s_i^{var}, s_i^{rule})$, $i = 1, \ldots, k - 1$, where $s_i^{pos}$, $s_i^{var}$ and $s_i^{rule}$ define the *position*, *predictor* and *rule* of each splitting node, respectively. For these parameters the priors can be specified as follows. First we can define a maximal number of splitting nodes, say, $s_{\max} = n - 1$, so $s_i^{pos} \in \{1, \ldots, s_{\max}\}$. Second we draw any of the $m$ predictors from an uniform discrete distribution $U(1, \ldots, m)$ and assign $s_i^{var} \in \{1, \ldots, m\}$. Finally the candidate value for the splitting variable $x_j = s_i^{var}$ is drawn from an uniform discrete distribution $U(x_j^{(1)}, \ldots, x_j^{(N)})$, where $N$ is the total number of possible splitting rules for predictor $x_j$, either categorical or continuous.

Such priors allow the exploring of DTs which partition data in as many ways as possible, and therefore we can assume that each DT with the same number of terminal nodes is equally likely [3]. For this case the prior for a complete DT is described as follows:

$$p(\boldsymbol{\theta}) = \left\{ \prod_{i=1}^{k-1} p(s_i^{rule} \mid s_i^{var}) p(s_i^{var}) \right\} p(\{s_i^{pos}\}_1^{k-1}). \qquad (3)$$

For a case when there is knowledge of the favoured structure of the DT, Chipman *et al.* [2] suggested a generalisation of the above prior – they assume the prior probability of further split of the terminal nodes to be dependent on how many splits have already been made above them. For example, for the $i$th terminal node the probability of its splitting is written as

$$p_{split}(i) = \gamma(1 + d_i)^{-\delta},$$ (4)

where $d_i$ is the number of splits made above $i$ and $\gamma$, $\delta \geq 0$ are given constants. The larger $\delta$, the more the prior favours "bushy" trees. For $\delta = 0$ each DT with the same number of terminal nodes appears with the same prior probability.

Having set the priors on the parameters $\phi$ and $\theta$, we can determine the marginal likelihood for the data given the classification tree. In the general case this likelihood can be written as a multinomial Dirichlet distribution [3]:

$$p(\mathbf{D} \mid \theta) = \left[ \frac{\Gamma\{\alpha C\}}{\{\Gamma(\alpha)\}^C} \right]^k \prod_{i=1}^{c} \frac{\prod_j^C \Gamma(m_{ij} + \alpha_j)}{\Gamma(n_i + \sum_{j=1}^{C} \alpha_j)},$$ (5)

where $n_i$ is the number of data points falling in the $i$th terminal node of which $m_{ij}$ points are of class $j$ and $\Gamma$ is a Gamma function.

To grow large DTs from real-world data, Chipman *et al.* [2] and Denison *et al.* [3] suggest exploring the posterior probability by using the following types of moves.

- *Birth*. Randomly split the data points falling in one of the terminal nodes by a new splitting node with the variable and rule drawn from the corresponding priors.
- *Death*. Randomly pick a splitting node with two terminal nodes and assign it to be one terminal with the united data points.
- *Change-split*. Randomly pick a splitting node and assign it a new splitting variable and rule drawn from the corresponding priors.
- *Change-rule*. Randomly pick a splitting node and assign it a new rule drawn from a given prior.

The first two moves, *birth* and *death*, are reversible and change the dimensionality of $\theta$ as described in [7]. The remaining moves provide jumps within the current dimensionality of $\theta$. Note that the *change-split* move is included to make "large" jumps which potentially increase the chance of sampling from a maximal posterior whilst the *change-rule* move does "local" jumps.

For the birth moves, the proposal ratio $R$ is written

$$R = \frac{q(\theta \mid \theta')p(\theta')}{q(\theta' \mid \theta)p(\theta)},$$ (6)

where the $q(\theta \mid \theta')$ and $q(\theta' \mid \theta)$ are the proposed distributions, $\theta'$ and $\theta$ are $(k + 1)$ and $k$-dimensional vectors of DT parameters, respectively, and $p(\theta)$ and $p(\theta')$ are the probabilities of the DT with parameters $\theta$ and $\theta'$:

$$p(\theta) = \{ \prod_{i=1}^{k-1} \frac{1}{N(s_i^{var})} \frac{1}{m} \} \frac{k}{S_k} \frac{1}{K},$$ (7)

where $N(s_i^{var})$ is the number of possible values of $s_i^{var}$ which can be assigned as a new splitting rule, $S_k$ is the number of ways of constructing a DT with $k$ terminal nodes, and $K$ is the maximal number of terminal nodes, $K = n - 1$.

For binary DTs, as given from graph theory, the number $S_k$ is the *Catalan number* [8] written as follows:

$$S_k = \frac{1}{k+1}\binom{2k}{k},$$ (8)

and we can see that for $k \geq 25$ this number becomes astronomically large, $S_k \geq (4.8)^{12}$.

The proposal distributions are as follows

$$q(\boldsymbol{\theta} \mid \boldsymbol{\theta}') = \frac{d_{k+1}}{D_Q'},$$ (9)

$$q(\boldsymbol{\theta}' \mid \boldsymbol{\theta}) = \frac{b_k}{k} \frac{1}{N(s_k^{var})} \frac{1}{m},$$ (10)

where $D_{Q1} = D_Q + 1$ is the number of splitting nodes whose branches are both terminal nodes.

Then the proposal ratio for a *birth* is given by

$$R = \frac{d_{k+1}}{b_k} \frac{k}{D_{Q1}} \frac{S_k}{S_{k+1}}.$$ (11)

The number $D_{Q1}$ in (11) is dependent on the DT structure and it is clear that $D_{Q1} < k \; \forall \; k = 1, \ldots, K$. Analysing (11), we can also assume that $d_{k+1} = b_k$. Then letting the DTs grow, i.e., $k \to K$, and considering $S_{k+1} > S_k$, we can see that the value of $R \to c$, where $c$ is a constant lying between 0 and 1.

Alternatively, for the death moves the proposal ratio is written as

$$R = \frac{b_k}{d_{k-1}} \frac{D_Q}{(k-1)} \frac{S_k}{S_{k-1}}.$$ (12)

We can see that under the assumptions considered for the birth moves, $R \geq 1$.

The DTs grow very quickly during the first burn-in samples because an increase in log likelihood value for the birth moves is much larger than that for the others. For this reason almost every new partition of data is accepted. Once a DT has grown the *change* moves are accepted with a very small probability and, as a result, the MCMC algorithm tends to get stuck at a particular DT structure instead of exploring all possible structures.

Because DTs are hierarchical structures, the changes at the nodes located at the upper levels can significantly change the location of data points at the lower levels. For this reason there is a very small probability of changing and then accepting a DT located near a root node. Therefore the MCMC algorithm collects the DTs in which the splitting nodes located far from a root node were changed. These nodes typically contain small numbers of data points. Subsequently, the value of log likelihood is not changed so much, and such moves are usually accepted. As a result, the MCMC algorithm cannot explore a full posterior distribution.

One way to extend the search space is to restrict DT sizes during a given number of the first burn-in samples as described in [3]. Indeed, under such a restriction, this strategy gives more chances of finding DTs of a smaller size which could be competitive in term of the log likelihood values with the larger DTs. The restricting strategy, however, requires setting up in an *ad hoc* manner the additional parameters such as the size DTs and the number of the first burn-in samples. Sadly, in practice, it often happens that after the limitation period the DTs grow quickly again and this strategy does not improve the performance.

Alternatively to the above approach based on the explicit limitation of DT size, the search space can be extended by using a *restarting strategy* as Chipman *et al.* have suggested in [2].

## 2.2. The Restarting Strategy

The idea behind the restarting strategy is based on multiple runs of the MCMC search algorithm with short intervals of burn-in and post burn-in. For each run, the MCMC creates an initial DT with the random parameters and then starts exploring the tree model space. Running short intervals prevents the DTs from getting stuck at a particular DT structure. More important, however, is that the multiple runs allow exploring of the DT model space starting with very different DTs. So, averaging the DTs over all such runs, we can improve the performance of the MCMC search algorithm.

The restarting strategy, as we see, does not limit the DT sizes explicitly as does the restricting strategy. For this reason the restarting strategy seems to be the more principled, and thus we use this strategy in our further experiments on comparing the performance with the randomised DT technique which is briefly described next.

## 3.    The Randomised Decision Tree Ensemble Technique

Performance of a single DT can be improved by averaging the outputs of an ensemble of DTs [2]. Improvement is achieved if most of the DTs can correctly classify the data points misclassified by a single DT. The required diversity of the classifier outcomes is thought to be achieved if the DTs involved in an ensemble are independently induced from data. To achieve the required independence, Dietterich has suggested randomising the DT splits [5]. In this technique the best, in terms of information gain, 20 partitions for any node are calculated and one of these is randomly selected with uniform probability. The class posterior probabilities are calculated for all the DTs involved in an ensemble and then averaged.

A pruning factor, specified as a minimal number of data points allowed to fall in the terminal nodes, can affect the ensemble performance. However, within the randomised DT technique, this effect is insignificant when pruning does not exceed 10% of the number of the training examples [5]. More significantly the pruning factor affects the average size of the DTs, and consequently it has to be set reasonably.

The number of the randomised DTs in the ensemble is dependent on the classification problem and assigned by an user in an *ad hoc* manner. This technique permits the user to evaluate the diversity of the ensemble by comparing the performances of the ensemble and that of the best DT on a predefined validation data subset. The required diversity is achieved if the DT ensemble outperforms the best single DTs involved in the ensemble. Therefore this ensemble technique requires the use of *n*-fold cross-validation [4, 5]. In our experiments described in section 5 we used the above randomised DT ensemble technique. For all the domain problems the ensembles consist of 200 DTs. To keep the size of the DT acceptable, the pruning factor is set to be dependent on the number of the training examples. In particular, its value is set to 30 for problems with many training examples; otherwise it is 5 (in all cases this is less than the 10% level). The performance of the randomised DT ensembles is evaluated on 5 folds for each problem.

## 4.    The Uncertainty Envelope Technique

In general, the MCSs described in the above sections 2 and 3 consist of classifiers trained independently. In such a case, we can naturally assume that the inconsistency of the classifiers on a given datum $\mathbf{x}$ is proportional to the uncertainty of the MCS. Let the value of class posterior probability $P(c_j|\mathbf{x})$ calculated for class $c_j$ be an average over the class posterior probability $P(c_j|K_i, \mathbf{x})$ given on classifier $K_i$:

$$P(c_j \mid \mathbf{x}) = \frac{1}{N} \sum_{i=1}^{N} P(c_j \mid K_i, \mathbf{x}), \tag{13}$$

where $N$ is the number of classifiers in the ensemble.

As classifiers $K_1$, …, $K_N$ are independent each other and their values $P(c_j|K_i, \mathbf{x})$ range between 0 and 1, the probability $P(c_j|\mathbf{x})$ can be approximated as follows

$$P(c_j \mid \mathrm{x}) \approx \frac{1}{N} \sum_{i=1}^{N} I(y_i, t_i \mid \mathbf{x}), \tag{14}$$

where $I(y_i, t_i)$ is the indicator function assigned to be 1 if the output $y_i$ of the *i*th classifier corresponds to target $t_i$, and 0 if it does not.

The larger number of classifiers, $N$, the smaller is error of the approximation (14). For example, when $N = 500$, the approximation error is equal to 1%, and when $N = 5000$, it becomes equal to 0.4%.

It is important to note that the right side of Eq. (14) can be considered as a *consistency* of the outcomes of MCS. Clearly, values of the consistency, $\gamma = \frac{1}{N} \sum_{i=1}^{N} I(y_i, t_i \mid \mathbf{x})$, lie between 1/C and 1.

Analysing Eq. (14), we can see that if all the classifiers are degenerate, i.e., $P(c_j|K_i, \mathbf{x}) \in \{0, 1\}$, then the values of $P(c_j|\mathbf{x})$ and $\gamma$ become equal. The outputs of classifiers can be equal to 0 or 1, for example, when the data points of two classes do not overlap. In other cases, the class posterior probabilities of classifiers range between 0 and 1, and the $P(c_j|\mathbf{x}) \approx \gamma$. So we can conclude that the classification confidence of an outcome is characterised by the consistency of the MCS calculated on a given datum. Clearly, the values of $\gamma$ are dependent on how representative the training data are, what classification scheme is used, how well the classifiers were trained within a classification scheme, how close the datum $\mathbf{x}$ is to the class boundaries, how the data are corrupted by noise, and so on.

Let us now consider a simple example of an MCS consisting of $N = 1000$ classifiers in which 2 classifiers give a conflicting classification on a given datum $\mathbf{x}$ to the other 998. Then consistency $\gamma = 1 - 2/1000 = 0.998$, and we can conclude that the MCS was trained well and/or the data point $\mathbf{x}$ lies far from the class boundaries. It is clear that for new datum appearing in some neighbourhood of the $\mathbf{x}$, the classification uncertainty as the probability of misclassification is expected to be $1 - \gamma = 1 - 0.998 = 0.002$. This inference is truthful for the neighbourhood within which the prior probabilities of classes remain the same. When the value of $\gamma$ is close to $\gamma_{min} = 1/C$, the classification uncertainty is highest and a datum $\mathbf{x}$ can be misclassified with a probability $1 - \gamma = 1 - 1/C$.

From the above consideration, we can assume that there is some value of consistency $\gamma_0$ for which the classification outcome is confident, that is the probability with which a given datum $\mathbf{x}$ could be misclassified is small enough to be acceptable. Given such a value, we can now specify the uncertainty of classification outcomes in statistical terms. The classification outcome is said to be *confident and correct*, when the probability of misclassification is acceptably small and $\gamma \geq \gamma_0$.

Additionally to the confident and correct output, we can specify a *confident but incorrect* output referring to a case when almost all the classifiers assign a datum $\mathbf{x}$ to a wrong class whilst $\gamma \geq \gamma_0$. Such outcomes tell us that the majority of the classifiers fail to classify a datum $\mathbf{x}$ correctly. The confident but incorrect outcomes can happen for different reasons, for example, the datum $\mathbf{x}$ could be mislabelled or corrupted, or the classifiers within a selected scheme cannot distinguish the data $\mathbf{x}$ properly.

The remaining cases for which $\gamma < \gamma_0$ are regarded as *uncertain classifications*. In such cases the classification outcomes cannot be accepted with a given confidence probability $\gamma_0$ and the MCS labels them as uncertain.

Fig 1 gives a graphical illustration for a simple two-class problem formed by two Gaussian $N(0, 1)$ and $N(2, 0.75)$ for variable $x$. As the class probability distributions are given, an optimal decision boundary can be easily calculated in this case. For a given confident consistency $\gamma_0$, the integration over the class posterior distribution gives boundaries B1 and B2 within which the outcomes of the MCS are assigned within the Uncertainty Envelope technique to be confident and correct (CC), confident but incorrect (CI) or uncertain (U). If a decision boundary within a selected classification scheme is not optimal, the classification error becomes higher than a minimal Bayes error. So, for the Bayesian classifier and a given consistency $\gamma_0$, the probabilities of CI and U outcomes on the given data are minimal as depicted in Fig. 1.

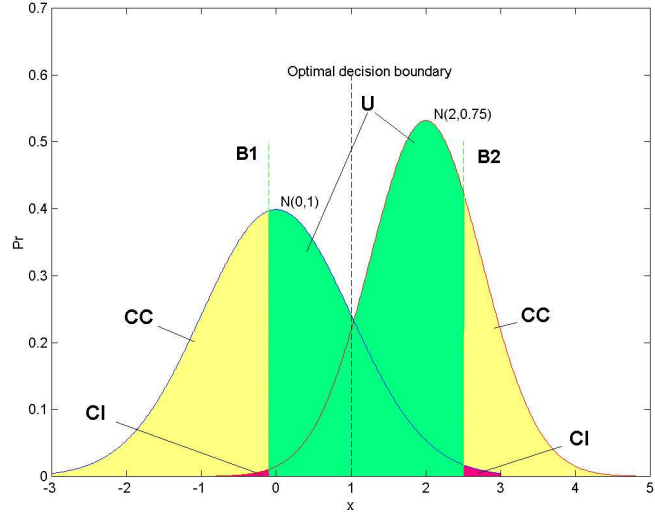

**Fig. 1:** Uncertainty Envelope characteristics for an example of two-class problem.

The above three characteristics, the confident and correct, confident but incorrect, and uncertain outcomes, seem to provide a practical way of evaluating different types of MCSs on the same data sets. Comparing the ratios of the data points assigned to be one of these three types of classification outcomes, we can quantitatively evaluate the classification uncertainty of the MCSs. Depending on the costs of types of misclassifications in real-world applications, the value of the confidence consistency $\gamma_0$ should be given, say, equal to 0.99.

Next, we compare the performance of the Bayesian and randomised DT ensembles within the described Uncertainty Envelope technique on some synthetic and real-world datasets.

## 5.    Experiments and Results

In this section we describe the results on the experimental comparison of two techniques, the Bayesian DT technique with the restarting strategy and the randomised DT ensemble technique. The experiments were conducted on a synthetic dataset, and then on some domain problems from the UCI Machine Learning Repository [9]. The performance of these MCSs is evaluated within the Uncertainty Envelope technique described in the above section.

### 5.1. Experiments with the Synthetic Data

In these experiments we use a two-dimensional synthetic problem, in order to visualise the decision boundaries and data points. The problem was generated by a mixture of five Gaussians. The data points drawn from the first three Gaussians belong to class 1 and the data points from the remaining two Gaussians to class 2. The mixing weights $\rho_{ij}$ and kernel centres $\mathbf{\mu}_{ij}$ of these Gaussians for class 1 and class 2 are

Class 1:

$\rho_{11} = 0.16$, $\mathbf{\mu}_{11} = (1.0, 1.0)$,

$\rho_{12} = 0.17$, $\mathbf{\mu}_{12} = (-0.7, 0.3)$,

$\rho_{13} = 0.17$, $\mathbf{\mu}_{13} = (0.3, 0.3)$.

Class 2:

$\rho_{21} = 0.25$, $\mathbf{\mu}_{21} = (-0.3, 0.7)$,

$\rho_{22} = 0.25$, $\mathbf{\mu}_{22} = (0.4, 0.7)$

All these kernels have isotropic covariance: $\mathbf{\Sigma}_i = 0.03\mathbf{I}$. Such a mixture of the kernels is an extended version of the Ripley data [10].

In our case the training data contain 250 and the test data 1000 data points drawn from the above mixture. Because the classes overlap, the Bayes error on the test data is 9.3%. The data points of the two classes denoted by the crosses and dots are depicted in Fig. 2. The Bayesian decision boundary is shown here by the dashed line. Both the Bayesian and randomised DT ensemble techniques were run on these synthetic data with the pruning factor $p_{min}$ set equal to 5.

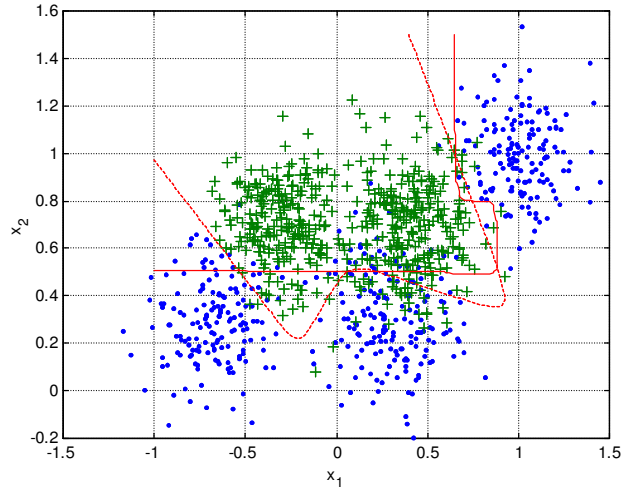

**Fig. 2:** Synthetic data points belonging to 2 classes. The dashed and solid lines depict the class boundaries calculated for the Bayesian rule and Bayesian DTs, respectively.

### 5.1.1. Performance of the Bayesian Decision Trees

The Bayesian DT technique with the restarting strategy was run 50 times; each time 2000 samples were taken for burn-in and 2000 for post burn-in. The probabilities of birth, death, change variable, and change rule were 0.1, 0.1, 0.1, and 0.7, respectively. The sample rate was set to 1. Priors on the number of nodes in DTs were set uniform. The uniform prior allows the DTs to grow by making birth moves while the proposed DT parameters made within the $p_{min}$ are available.

Fig. 3 depicts the samples drawn from the log likelihood calculated for DTs accepted during the burn-in and post burn-in phases for all the 50 runs. The total number of samples was $10^5$. From this figure we can see that during the burn-in phase the values of log likelihood quickly converge to a stable value at about –40, and during the post burn-in they randomly fluctuate around this value. This means that the MCMC stochastic sampling works well. The acceptance rates during the burn-in and post burn-in phases were equal to 0.47.

The middle plots in Fig. 3 depict the samples of DT size drawn during the burn-in and post burn-in. The bottom plots depict the distributions of the DTs over the number of DT nodes. The mean and variance values of DT nodes were 12.4 and 2.5, respectively.

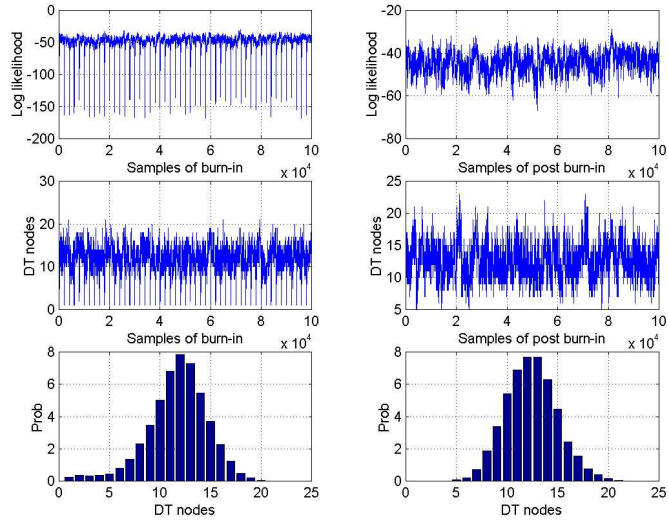

**Fig. 3:** Synthetic data: Samples of log likelihood and DT size during burn-in and post burn-in. The bottom plots are the distributions of DT sizes.

Diversity of the classifiers is one of the important characteristics determining the quality of Bayesian averaging over classification models. The diversity of the Bayesian DTs can be presented by the posterior distribution of the DTs accepted during post burn-in. The top plot in Fig. 4 shows such a distribution calculated in our experiments on the synthetic data.

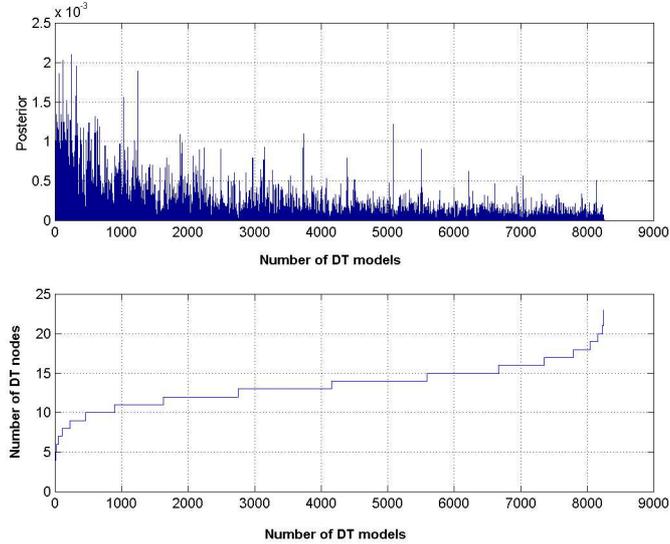

**Fig 4:** Synthetic data: Distribution of the accepted DTs sorted out on the number of nodes.

As we can see from the above distribution, the diversity of the DTs is very large because their number is more than 8000. These DT models were sorted on the number of splitting nodes. At the bottom plot of Fig. 4 we can see that the number of nodes is monotonically increasing from 6 up to 23. Analysing the distribution we can observe that the posterior weights of the DTs decrease when the number of splitting nodes increase. That is, the Bayesian MCMC technique explored the smaller DTs more frequently.

Table 1 lists the parameters of the first 20 DT models accepted during post burn-in with the highest posterior weights. These parameters are the path of features used for partitions beginning with a DT root and the number of nodes in DT. We can see that a DT which involves the features in an order 2-1-1-1-2-2-1-1-1 has a maximal posterior weight.

The resultant classification accuracy of the Bayesian DTs was 87.6% with $2\sigma$ interval equal 0.6%. The rates of confident and correct, uncertain and confident but incorrect outcomes were 63.3%, 34.4% and 2.3%, respectively. The $2\sigma$ intervals for these estimates had widths 15.7%, 20.1%, and 2.9% respectively.

The decision boundary averaged over the Bayesian DTs is shown by the solid line in Fig. 2 above. We can see that there are some regions in which the average decision boundary does not fit to the data as well as the Bayesian rule depicted by the dashed line.

**Table 1:** The first 20 DT models accepted during post burn-in with the highest posterior weights.

| # | Path of features | Number of nodes in DT | Posterior weights |
|---|---|---|---|
| 1 | 211122111 | 9 | 0.002 |
| 2 | 21112211 | 8 | 0.002 |
| 3 | 212212111 | 9 | 0.002 |
| 4 | 21221211111 | 11 | 0.002 |
| 5 | 2112222 | 7 | 0.002 |
| 6 | 212211111 | 9 | 0.002 |
| 7 | 21121212222 | 11 | 0.002 |
| 8 | 22121121 | 8 | 0.002 |
| 9 | 211122112 | 9 | 0.001 |
| 10 | 22121112 | 8 | 0.001 |
| 11 | 2112221 | 7 | 0.001 |
| 12 | 211222 | 6 | 0.001 |
| 13 | 22211111 | 8 | 0.001 |
| 14 | 2212211 | 7 | 0.001 |
| 15 | 2121112 | 7 | 0.001 |
| 16 | 2121212112 | 10 | 0.001 |
| 17 | 2122121111 | 10 | 0.001 |
| 18 | 211112121 | 9 | 0.001 |
| 19 | 21122222 | 8 | 0.001 |
| 20 | 2222111 | 7 | 0.001 |

### 5.1.2. Performance of the Randomised Decision Tree Ensemble

On the synthetic data, the ensemble output quickly converges and stabilizes after averaging approximately. 100 DTs. As an example, Fig. 4 depicts the convergence of the ensemble outputs over the 5 folds.

During the averaging, the ensemble output converges to a stable value quickly. Fig. 6, for example, depicts the performances of the ensemble, single and the best DT selected on the validation subset on the 5th fold. In this figure, the bold line marked Pe is the performance of the DTs averaged within the ensemble, the thin line marked Ps is the performance of a single DT, and the dashed line marked Pslv is the performance of the best DT on the validation subset. As we can see, the ensemble performance, Pe, becomes stable after averaging 130 DTs and its value stays higher than that of the best DT, Pslv, selected on the validation subset.

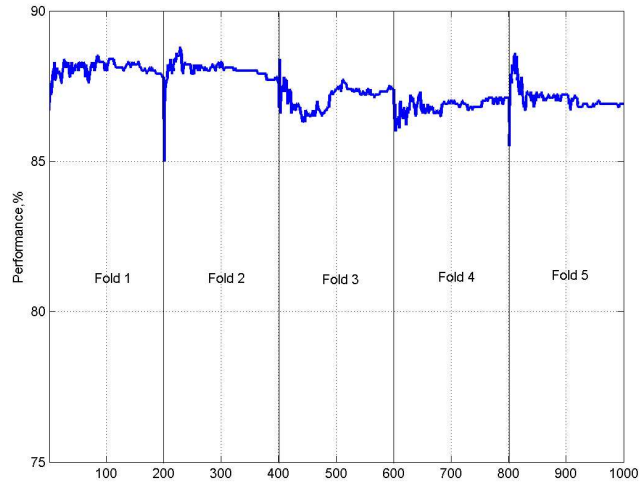

**Fig 5:** Synthetic data: Performance of the DT ensemble over 5 folds.

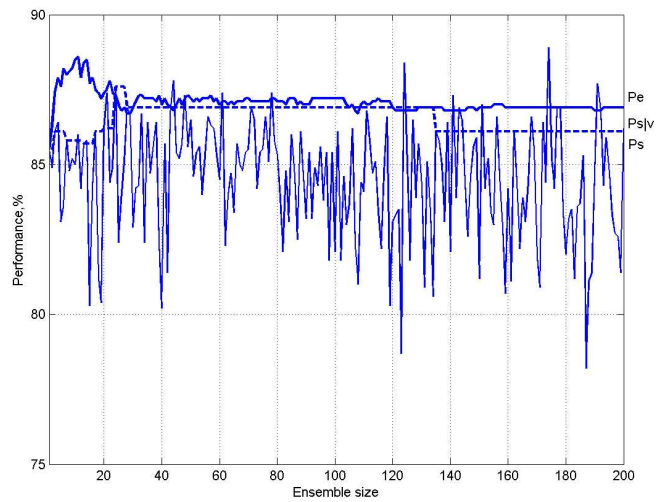

**Fig 6:** Synthetic data: Performances of the randomised DT ensemble, as well a single DT and the best DT selected on the validation subset on the 5th fold.

The distribution of DTs over the number of splitting nodes is shown in Fig. 7. This distribution was calculated on all 5 folds. The average size of DTs and the standard deviation were 32.9 and 3.3, respectively.

The averaged classification performance was 87.1%. Within the Uncertainty Envelope, the rates of confident and correct, uncertain, and confident but incorrect outcomes were 78.9%, 9.8%, and 11.3%, respectively. The widths of 2σ intervals for these estimates were 34.9%, 43.7%, and 8.9%, respectively. We can see that the values of these intervals are very large. This happens because the randomised DT ensemble technique gives the outcomes with a very high rate of uncertain classifications on some of the folds. In other words this technique is not stable enough.

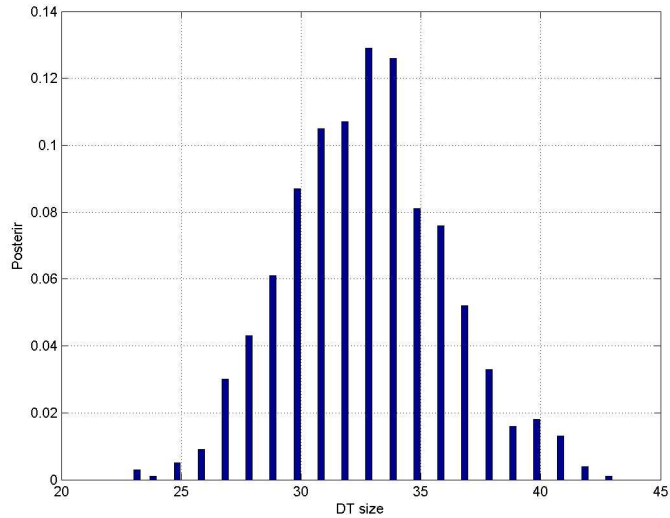

**Fig 7:** Synthetic data: Distribution of DTs over the number of splitting nodes over the 5 folds.

### 5.1.3. Comparison of Performances

Comparing the experimental results on the synthetic data, we can see that both techniques provide the same performance in terms of the classification accuracy on the test data. However, the size of Bayesian DTs is, on average, in one-third the size of the randomised ensemble. Meanwhile within an Uncertainty Envelope technique, the Bayesian averaging over DTs provides more reliable estimates of the classification uncertainty than the averaging over the randomised DTs: as we can see, the variances of the confident and correct and confident but incorrect outcomes calculated by the Bayesian model averaging technique are significantly less than those calculated by the ensemble averaging technique: 15.7%, 20.1%, and 2.9% versus 34.9%, 43.7%, and 8.9%, respectively. In other words, the Bayesian DT technique provides more stable classification outcomes than the randomised DT ensemble technique.

### 5.2. Experiments with the UCI Machine Learning Depository Datasets

In these experiments we used the 7 domain problems taken from the UCI Machine Learning Repository [9]. Table 2 lists the names and characteristics of these problems, here *C*, *m*, *train*, and *test* are the numbers of classes, the number of input variables, the number of training and test examples, respectively. This table also provides the performances of the best single DTs on the validation datasets.

The performances of the randomised DT ensemble technique within the Uncertainty Envelope technique are shown in Table 3. From this table we can see first that the randomised DT ensembles always outperform the best single DTs. Second the $2\sigma$ intervals calculated for the confident correct and incorrect outcomes are very large on the Ionosphere, Image, Sonar, Vehicle, and Pima problems.

**Table 2:** The data characteristics and performance of the best single DTs

| Data | Data characteristics | | | | Perform, % |
|---|---|---|---|---|---|
| | C | M | train | test | |
| Ionosphere | 2 | 33 | 200 | 151 | 88.8±8.0 |
| Wisconsin | 2 | 9 | 455 | 228 | 96.1±1.7 |
| Image | 7 | 19 | 210 | 2100 | 87.4±4.4 |
| Votes | 2 | 16 | 391 | 44 | 93.9±3.1 |
| Sonar | 2 | 60 | 138 | 70 | 70.7±7.8 |
| Vehicle | 4 | 18 | 564 | 282 | 69.0±4.5 |
| Pima | 2 | 8 | 512 | 256 | 77.3±1.2 |

**Table 3:** Performances of the randomised DT ensembles within the Uncertainty Envelope

| Data | DT size | Perform % | Uncertainty Envelope, % | | |
|---|---|---|---|---|---|
| | | | Correct | Uncertain | Incorrect |
| Ionosphere | 21.2±1.3 | 94.4±0.7 | 76.5±35.8 | 7.0±44.4 | 16.5±18.4 |
| Wisconsin | 32.7±1.5 | 97.7±1.2 | 96.7±7.9 | 1.4±9.2 | 1.9±1.8 |
| Image | 27.9±1.3 | 94.2±0.9 | 86.1±33.0 | 6.5±37.9 | 7.4±7.9 |
| Votes | 27.1±3.6 | 95.2±1.4 | 94.3±5.8 | 1.1±7.2 | 4.5±2.1 |
| Sonar | 17.8±0.8 | 78.3±5.5 | 54.9±40.6 | 9.6±60.5 | 35.6±31.8 |
| Vehicle | 115.8±3.2 | 71.9±2.2 | 63.8±31.0 | 8.8±50.2 | 27.4±20.1 |
| Pima | 33.6±4.0 | 80.2±2.4 | 66.7±47.0 | 14.6±65.3 | 18.7±19.6 |

Likewise, Table 4 lists the performances of the Bayesian DTs. From this table we can see that first the performances in terms of classification accuracy on the test data are nearly the same excluding the Wisconsin and Sonar problems on which the Bayesian DT ensembles slightly out-perform the randomised DT ensembles. The Bayesian DTs are smaller on average than those of the randomised ensemble by a factor of 2.4.

**Table 4:** Performances of the Bayesian DTs with a restarting strategy within the Uncertainty Envelope

| Data | DT size | Perform, % | Uncertainty Envelope, % | | |
|---|---|---|---|---|---|
| | | | Correct | Uncertain | Incorrect |
| Ionosphere | 12.8±3.2 | 95.3±0.6 | **12.1±5.2** | 87.3±6.8 | **0.6**±0.8 |
| Wisconsin | 12.4±1.4 | **99.1**±0.8 | 81.3±2.7 | 18.3±4.4 | 0.3±0.7 |
| Image | 14.9±2.8 | 94.3±0.3 | **23.3±1.4** | 76.6±4.9 | **0.0**±0.0 |
| Votes | 12.0±2.1 | 95.4±1.2 | 53.5±12.3 | 44.0±12.5 | **2.5**±2.2 |
| Sonar | 10.2±1.9 | **81.4**±3.1 | **2.2±1.0** | 97.8±2.7 | **0.0**±0.0 |
| Vehicle | 45.3±3.9 | 69.9±3.5 | **2.9±1.7** | 96.9±1.3 | **0.2**±0.5 |
| Pima | 12.2±2.0 | 79.7±1.7 | **33.5±6.3** | 62.1±6.5 | **4.4**±1.9 |

From Table 4 we can see that the variances of the confident and correct as well as the confident but incorrect outcomes calculated by the Bayesian DTs are significantly less than those calculated by the randomised DT ensembles. This means that the Bayesian DT technique is capable of providing more stable classification outcomes than the randomised DT ensemble technique.

Our special interest in comparing Tables 3 and 4 is the discrepancy between the sizes of the respective "uncertain" classification rates for the two DT techniques. For a given confident probability 0.99, both techniques provide rather similar average classification accuracy, however the Bayesian DTs provide more narrow intervals around the average values than the randomised DTs.

Clearly, adjusting the confidence probabilities can give us additional information about comparison of the classification uncertainty of the two techniques. In particularly, gradually increasing the confident probability in our experiments on the Wisconsin Data, the classification uncertainty rates are changed as depicted by Fig. 8.

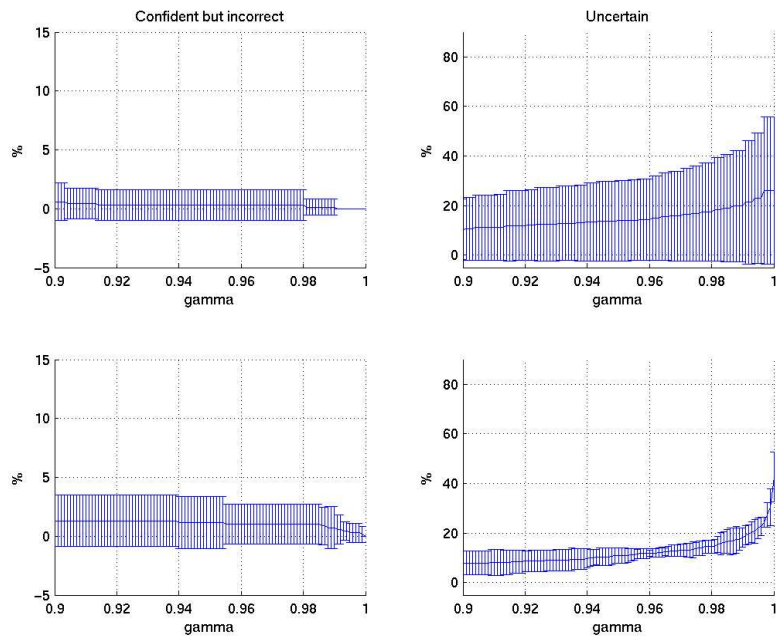

**Fig 8:** Wisconsin Data: The classification uncertainty rates *versus* the values of confident probabilities *gamma* for the randomised (the top plots) and Bayesian (the bottom plots) DT techniques.

In Fig. 8, the top plots present the error bars calculated for estimates of confident but incorrect and uncertain classifications made by the randomized DTs *versus* values of confident probability, *gamma*, increasing from 0.9 to 1.0 with step 0.001. Likewise, the bottom plots present the error bars calculated for the confident but incorrect and uncertain classifications made by the Bayesian DTs. These bars were calculated for $2\sigma$ intervals within the 5 fold cross-validation.

Comparing the above results, we can see that both techniques provide rather similar averaged rates calculated for confident but incorrect classifications as well as for uncertain classifications. However, comparing the $2\sigma$ intervals calculated for uncertain classifications, we can see that the Bayesian DT technique provides significantly better results than the randomized technique.

Because the computational cost required to implement the randomized as well as the Bayesian DT ensemble techniques is high, there are some open questions which are currently under the investigation. We hope to obtain the new results in order to give the exhaustive answers on these questions in the future work.

## 6.    Conclusion

We have experimentally compared the classification uncertainty of the Bayesian DT technique sampling posterior using MCMC with a restarting strategy and the randomised DT ensemble technique on an artificial data as well as on the Machine Learning Repository problems. The ensemble techniques both outperform the best single DTs and have rather similar average classification accuracy on the test datasets. However, the Bayesian ensembles make far fewer confident but incorrect classifications. This is clearly a very desirable property for multiple classifier systems applied to safety-critical problems for which confidently made, but incorrect, classifications may be fatal.

### Acknowledgments

This research was supported by the EPSRC, grant GR/R24357/01.